\documentclass[runningheads]{llncs}
\usepackage[T1]{fontenc}
\usepackage[T1]{fontenc}
\usepackage{mathtools} 
\usepackage{amsmath}
\usepackage{amsfonts}
\usepackage{multirow}
\usepackage{amssymb}
\usepackage{mathtools}
\usepackage{blindtext}
\usepackage{booktabs}
\usepackage{subcaption}
\usepackage{bm}
\usepackage[dvipsnames]{xcolor}
\usepackage{tikz}
\usepackage{hyperref}
\usepackage{fontawesome}
\usepackage{enumitem}
\usepackage{orcidlink}
\hypersetup{
    colorlinks=true,
    linkcolor=blue,
    filecolor=magenta,      
    urlcolor=cyan,
    pdftitle={Overleaf Example},
    pdfpagemode=FullScreen,
    }
\hypersetup{
    colorlinks=true,
    linkcolor=blue,
    filecolor=magenta,      
    urlcolor=cyan,
    pdftitle={Overleaf Example},
    pdfpagemode=FullScreen,
    }
\usepackage{graphicx}
\usepackage{graphicx,verbatim}
\begin{document}
%
%\title{Probabilistic Embedding Informed Coresets for Robust Visual In-Context Learning}
\title{Geometry-Aware Uncertainty Coresets for Robust Visual In-Context Learning in Histopathology}

\titlerunning{Geometry-Aware Uncertainty Coresets}
\begin{comment}  %% Removed for anonymized MICCAI submission
\author{First Author\inst{1}\orcidID{0000-1111-2222-3333} \and
Second Author\inst{2,3}\orcidID{1111-2222-3333-4444} \and
Third Author\inst{3}\orcidID{2222--3333-4444-5555}}
%
\authorrunning{F. Author et al.}
\institute{Princeton University, Princeton NJ 08544, USA \and
Springer Heidelberg, Tiergartenstr. 17, 69121 Heidelberg, Germany
\email{lncs@springer.com}\\
\url{http://www.springer.com/gp/computer-science/lncs} \and
ABC Institute, Rupert-Karls-University Heidelberg, Heidelberg, Germany\\
\email{\{abc,lncs\}@uni-heidelberg.de}}

\end{comment}
\author{Franciskus Xaverius Erick\inst{1}\orcidlink{0000-0001-6004-7896} \and Johanna Paula Müller\inst{1}\orcidlink{0000-0001-8636-7986}  \and \\ Bernhard Kainz\inst{1,2}\orcidlink{0000-0002-7813-5023}}
%Franciskus Xaverius Erick\inst{1}\orcidID{0000-1111-2222-3333} \and
%Second Author\inst{2,3}\orcidID{1111-2222-3333-4444} \and
%Bernhard Kainz\inst{3}\orcidID{2222--3333-4444-5555}}
\institute{FAU Erlangen-Nürnberg, Erlangen, DE \\\email{franciskus.erick@fau.de} \and Department of Computing, Imperial College London, London, UK}%\\
  
\maketitle              % typeset the header of the contribution
\begin{abstract}

Vision-language models (VLMs) can couple visual perception with open-ended clinical reasoning, making them attractive for computational histopathology. However, fine-tuning billions of parameters on scarce, expert-annotated pathology data is prohibitive, while in-context learning (ICL), which conditions the VLM on demonstrative image-text pairs without parameter updates, suffers from high sensitivity to which examples are selected and how the query is phrased, producing unreliable diagnostics. Existing selection strategies rely on query-dependent nearest-neighbour retrieval that ignores global data structure, require costly parameter updates, or disregard the joint vision-text embedding geometry of VLMs. We propose GAUC, a training-free coreset selection method operating directly in the pre-trained multimodal embedding space. GAUC jointly optimises three objectives: (1) a Maximum Mean Discrepancy term enforcing distributional fidelity between coreset and full dataset, (2) an Effective Mutual Information Difference regulariser bounding performance degradation under prompt paraphrases by exploiting the VLM's joint vision-text alignment, and (3) a predictive-uncertainty (entropy) penalty suppressing ambivalent, hallucination-prone outputs. On CRC-100K and MHIST across multiple open-source VLM architectures, GAUC \emph{matches} the accuracy of the strongest ICL selection and dataset-distillation baselines while \emph{substantially} improving calibration, prompt robustness, and hallucination rates, all without a single gradient update. Code is available at \href{https://anonymous.4open.science/r/submission-03BD}{\faGithub\ Anonymous Repository}
\keywords{Vision-Language Model \and Coreset Selection \and In-Context Learning \and Uncertainty.}
\

\iffalse
Training vision language models for medical imaging diagnosis purposes pose a considerable challenge due to the large computational requirements and labelled samples required for the training of billions of VLM model parameters. One possible solution is to leverage the in-context learning capability of VLMs, whereby demonstrative examples are provided in the query. However, in-context learning is highly dependent on the in-context examples chosen, leading to high variance in performance. The methods presented in the literature are either sensitive to the encoder model used or are prohibitively expensive. We propose a simple yet mathematically robust example coreset selections which aim to leverage the embedding knowledge of the VLMs themselves without further training of the pre-trained parameters. 
Our framework consists of the following components: 1) Geometry-aware coreset selection based on the embedding representations of the VLM encoders used, thereby decreasing variability between VLM models used, 2) an additional probabilistic embedding shift regularisation term, which minimises the variance induced by small variations of query texts, 3) an entropy minimisation coreset regularisation, enabling uncertainty-aware extraction of representative example coresets. We evaluate our method with the CRC100K and MHIST dataset.
\keywords{Vision Language Model  \and Coreset Selection \and Uncertainty.}
% Authors must provide keywords and are not allowed to remove this Keyword section.
\fi
    %your solution, contribution

\end{abstract}
\section{Introduction}
Histopathological examination remains the diagnostic gold standard for most solid tumours, requiring trained pathologists to visually assess tissue morphology at high magnification, a process that is time-consuming, subjective, and bottlenecked by a global shortage of specialist expertise~\cite{campanella2019clinical}. With colorectal cancer alone accounting for over 1.9~million new cases and 935,000 deaths annually~\cite{ferlay2021global}, scalable tools that support high-throughput tissue classification are  needed. The widespread adoption of automated sample preparation pipelines and whole-slide image (WSI) scanners has further intensified this demand by dramatically accelerating the rate at which digitised histology data is generated, placing mounting workload pressure on pathologists~\cite{campanella2019clinical}. Pre-trained vision-language models (VLMs) offer a compelling path forward: by coupling a vision encoder with a large language backbone, they interpret complex medical imagery while producing human-readable diagnostic rationales that mirror clinician reasoning~\cite{li2023llavamed,kurz2025benchmarking}. Unlike conventional supervised classifiers, VLMs can simultaneously process visual evidence and contextual clinical descriptions, enabling more holistic and interpretable tissue analysis~\cite{ferber2024incontext}.

However, deploying VLMs clinically is hindered by the intractability of fine-tuning billions of parameters on scarce, privacy-sensitive data~\cite{kurz2025benchmarking} and by persistent overconfident hallucinations that are unacceptable in safety-critical settings~\cite{cheng2025mitigating,jaeger2023understanding}. In-context learning (ICL) sidesteps the first obstacle by conditioning the model on demonstrative image-label pairs placed directly in the prompt, requiring no parameter updates~\cite{brown2020languagemodelsfewshotlearners}. This capability, originally observed in large language models, transfers naturally to VLMs~\cite{openai2024gpt4technicalreport,awadalla2023openflamingoopensourceframeworktraining,laurencon2024mattersbuildingvisionlanguagemodels}. Yet ICL is notoriously sensitive to both the choice of demonstrations and the phrasing of the textual query~\cite{lu2022fantastically,zhao2021calibrateuseimprovingfewshot,li2023configure}. Current remedies either insert learnable shift modules that require additional training~\cite{jiang2025mimicincontextlearningmultimodal,li2023configure}, or rely on nearest-neighbour retrieval that captures local query similarity but ignores the global distributional structure of the dataset~\cite{ferber2024incontext}. Neither accounts for the joint visual-textual embedding geometry of the VLM, nor explicitly controls for prompt-induced instability or predictive overconfidence, shortcomings that are especially damaging in histopathology where class imbalance obscures minority morphologies and minor prompt reformulations can flip a diagnosis. 
We propose a principled, entirely training-free coreset selection framework for visual ICL that jointly optimises representativeness, prompt robustness, and predictive calibration. Our contributions are:
\begin{enumerate}[leftmargin=*,nosep]
\item A geometry-aware objective based on Maximum Mean Discrepancy (MMD) that selects real, clinically traceable demonstrations whose embedding distribution preserves the global structure of the full dataset, preventing collapse into dominant morphological clusters.
\item An Effective Mutual Information Difference (EMID)-derived mutual-information regulariser that quantifies response discrepancies under paraphrased queries, steering selection toward in-context sets robust to textual variation across both visual and textual modality embeddings.
\item A variance regularisation term that penalises predictive uncertainty, encouraging demonstrations that suppress overconfident hallucinations.
\item Consistent improvements over nearest-neighbour and random baselines on two challenging histopathology benchmarks, CRC-100K (8-class tissue subtyping) and MHIST (binary polyp classification), with fully training-free deployment compatible with any off-the-shelf VLM. Reliability, not accuracy, is the central contribution.
\end{enumerate}

\noindent\textbf{Related Work.}
Recent works in ICL~\cite{liu2024incontext} have formalised demonstrations as latent shift vectors that steer query-token representations, and extended this view to multimodal models through lightweight query-dependent shift modules~\cite{jiang2025mimicincontextlearningmultimodal} and prompt-configuration strategies~\cite{li2023configure}. These methods are effective but require learnable parameters or task-specific inference modifications.
In computational pathology, $k$-Nearest Neighbour ($k$NN) retrieval of demonstration images has matched or exceeded specialised fine-tuned networks~\cite{ferber2024incontext}, yet $k$NN selects solely by local query proximity. Two broader strategies exist for constructing representative subsets: dataset distillation synthesises artificial samples by matching gradient trajectories, feature distributions, or diffusion-based latent mappings~\cite{zhao2025tamingdiffusiondatasetdistillation,cechnicka2023realistic,cechnicka2024urcdm} but is computationally expensive and produces clinically untraceable images, whereas coreset selection retrieves real samples and can enforce distributional fidelity through statistical distances such as Maximum Mean Discrepancy~\cite{wang2025efficient}.
Curriculum-based adaptation on biomedical figure-caption corpora effectively aligns VLM semantics with visual features~\cite{li2023llavamed}, yet clinical VLMs remain prone to overconfident hallucinations and prompt-induced output instability~\cite{cheng2025mitigating,jaeger2023understanding,kurz2025benchmarking}. The Effective Mutual Information Difference (EMID) upper-bounds the performance degradation from multimodal alignment shifts via Jensen-Shannon divergence in the joint latent space~\cite{oh2025understanding}.

%\noindent\textbf{Related Work:}...

\section{Method}

\begin{figure}[t]
    \centering
    \includegraphics[width=\linewidth]{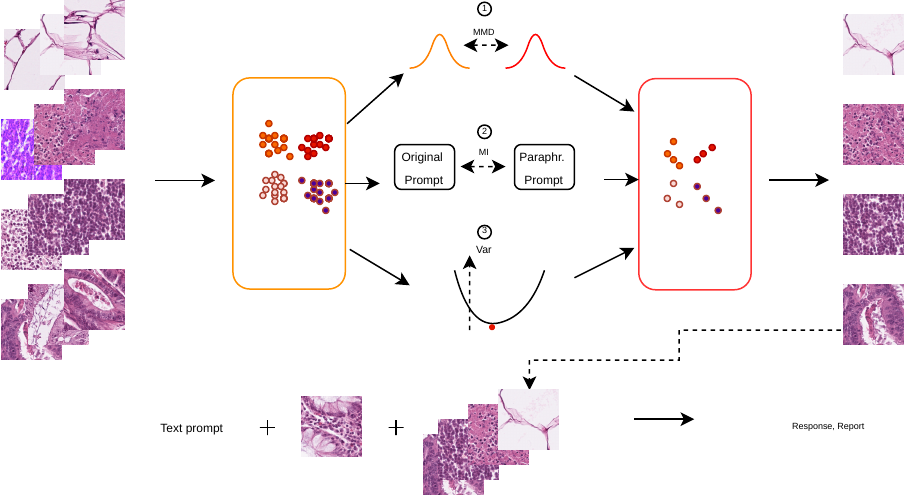}
    \caption{The GAUC optimization pipeline. GAUC involves 1) MMD matching between the embedded distribution of the full dataset and the selected coreset examples, 2) effective mutual information upper bound regularisation between outputs with original text prompts and paraphrased prompts, and 3) variance uncertainty minimisation of selected coreset examples. }
    \label{fig:pipeline}
\end{figure}

Let $W_\theta$ denote a pre-trained VLM with frozen parameters $\theta$. Given a full labelled dataset $F = \{(x_i, y_i)\}_{i=1}^{|F|}$ of histopathology patches and their class labels, we seek a compact coreset $\mathcal{D} = \{(x_j, y_j)\}_{j=1}^{|\mathcal{D}|}$ with $|\mathcal{D}| \ll |F|$ that, when placed in the prompt as in-context demonstrations, maximises diagnostic accuracy while remaining robust to prompt variation and predictive overconfidence. We optimise $\mathcal{D}$ by minimising a composite objective over three complementary terms (Fig.~\ref{fig:pipeline}), each operating entirely in the pre-trained embedding space of $W_\theta$ without any gradient-based parameter updates.

\noindent\textbf{Geometry-aware coreset selection via MMD.}
We require the embedding distribution of $\mathcal{D}$ to faithfully mirror that of $F$. Let $x, x'$ denote independent samples from $F$ and $u, u'$ from $\mathcal{D}$, mapped through the vision encoder of $W_\theta$. We measure distributional discrepancy with the Maximum Mean Discrepancy in a reproducing kernel Hilbert space~\cite{tolstikhin2016}:
\begin{equation}\label{eq:mmd}
\mathrm{MMD}^2(F, \mathcal{D})
=
\mathbb{E}_{x,x'}\!\big[k(x,x')\big]
+
\mathbb{E}_{u,u'}\!\big[k(u,u')\big]
-
2\,\mathbb{E}_{x,u}\!\big[k(x,u)\big],
\end{equation}
where $k(x,u) = \exp\!\bigl(-\|x - u\|^2 / 2\sigma^2\bigr)$ is the Gaussian (RBF) kernel, chosen because it is \emph{characteristic}: the induced MMD vanishes iff the two distributions coincide~\cite{gretton2012kernel}, so driving $\mathrm{MMD}^2\!\to\!0$ matches the full embedding distribution rather than only its low-order moments (as a linear kernel would). We set the bandwidth by the median heuristic, $\sigma^2 = \operatorname{median}\{\|x_i - x_j\|^2\}$. Minimising $\mathrm{MMD}^2$ prevents the coreset from collapsing into locally dominant clusters and ensures global representativeness across tissue morphologies.

\noindent\textbf{Prompt-robustness regularisation via EMID.}
Minor textual rephrasing of the query prompt can shift the conditional response distribution of a VLM, degrading diagnostic reliability. We regularise against this instability using EMID~\cite{oh2025understanding}. For a query image $x$, coreset $\mathcal{D}$, original prompt $t$, and paraphrased variant $t'$, the mutual information between the model response $r$ and the multimodal input is $I(r;\, x, \mathcal{D}, t) = H(r) - H(r \mid x, \mathcal{D}, t)$. The EMID quantifies how this coupling degrades under the prompt shift $t \to t'$: $\operatorname{EMID}(P,Q) = \operatorname{EMI}\!\bigl(P(r \mid x, \mathcal{D}, t)\bigr) - \operatorname{EMI}\!\bigl(Q(r' \mid x, \mathcal{D}, t')\bigr)$, where $P$ and $Q$ denote the response distributions under the original and paraphrased prompts. A tractable upper bound decomposes into Jensen-Shannon divergences over the individual modality embeddings:
\begin{equation}\label{eq:emid}
\operatorname{EMID}_{\text{upper}} =
D_{JS}^{1/2}(P_x \| Q_x)
+ D_{JS}^{1/2}(P_t \| Q_t)
+ D_{JS}^{1/4}(P_{\hat{r}} \| P_r)
+ D_{JS}^{1/4}(P_{\hat{r}} \| Q_{r'}),
\end{equation}
where $P_x, Q_x$ and $P_t, Q_t$ are the visual and textual embedding distributions under each prompt, $P_r, Q_{r'}$ the corresponding response distributions, and $\hat{r}$ the ideal ground-truth response. We treat prompt paraphrases, generated by a separate LLM, as localised distribution shifts in the textual modality and penalise coresets that yield large $\operatorname{EMID}_{\text{upper}}$. Unlike retrieval methods that operate solely on image embeddings~\cite{wang2025efficient}, this term explicitly leverages the joint vision-text alignment of the VLM to enforce prompt invariance.

\noindent\textbf{Predictive uncertainty regularisation.}
% TODO (Erick): confirm this matches the implemented quantity. The submitted
% definition used Var over class labels, for which "minimise -> concentrated"
% is directionally inconsistent (a peaked posterior has HIGH across-class
% variance). We reframe the term as predictive entropy, for which minimisation
% correctly yields concentrated predictions and matches the hallucination
% narrative in Table 3. If your code computes a different quantity, adjust here;
% consider also renaming the symbol from Var to a dedicated symbol for clarity.
To discourage demonstration sets that leave the model in ambivalent, high-uncertainty states---empirically the regime most prone to hallucinated findings (Table~\ref{tab:robust})---we penalise the predictive entropy of the class-posterior, $\operatorname{Var}(\mathcal{D}) = -\sum_{k \in \mathcal{Y}} p(y_k \mid x, \mathcal{D})\log p(y_k \mid x, \mathcal{D})$ (we retain the symbol $\operatorname{Var}(\mathcal{D})$ for consistency with Eq.~\eqref{eq:objective} and Table~\ref{tab:ablation}). Minimising this entropy steers selection toward concentrated predictions; representativeness (MMD) and prompt-invariance (EMID) guard against confident \emph{errors}, while this term removes residual ambivalence.

\noindent\textbf{Joint objective.}
The final coreset is obtained by jointly minimising:
\begin{equation}\label{eq:objective}
\mathcal{D}^* = \arg\min_{\mathcal{D}}
\;\mathrm{MMD}^2(F, \mathcal{D})
\;+\;
\alpha\,\operatorname{EMID}_{\text{upper}}
\;+\;
\beta\,\operatorname{Var}(\mathcal{D}),
\end{equation}
with $\alpha, \beta \geq 0$ controlling the trade-off between distributional fidelity, prompt robustness, and predictive stability.  Each term upper-bounds a distinct source of unreliability under a stated assumption; the formal theoretical justifications and proofs are in the supplementary material. Because all three terms are evaluated from forward-pass embeddings and output log-probabilities, the entire optimisation is training-free, requires no backward passes through $W_\theta$, and produces a single query-independent coreset reusable across all test images.

\section{Experiments}\label{sec:experiments}

We evaluate on two histopathology benchmarks. \textbf{CRC-100K}~\cite{kather2019} contains 100,000 H\&E-stained colorectal tissue patches spanning 9 classes; following prior work \cite{ferber2024incontext,Filiot2023,wang2021miccai} we omit the background class and report 8-class performance. \textbf{MHIST} \cite{wei2021petri} consists of 3,152 colorectal polyp patches annotated as hyperplastic polyp (HP) or sessile serrated adenoma (SSA), a binary task that is challenging even for trained pathologists. All experiments use two open-source VLM families, Qwen and LLaVA, loaded with default ImageNet pre-trained weights from HuggingFace to test out-of-domain ICL capability. We optimise coresets via greedy selection for 1,000 iterations. Since the three terms have different scales, we standardise each by its standard deviation over the candidate pool so that $\alpha,\beta$ are relative weights, then select them by grid search on a held-out validation split (minimising validation ECE subject to accuracy within $1\sigma$ of the best). This yields $\alpha=\beta=0.1$, within a broad stable plateau (see supplementary).

\noindent\textbf{Baselines.}
We compare against three ICL demonstration-selection strategies: random sampling, $k$NN retrieval~\cite{ferber2024incontext}, and mutual-information-informed retrieval (DR)~\cite{wang2025efficient}, as well as the shift-vector method MIMIC~\cite{jiang2025mimicincontextlearningmultimodal} which requires additional training. We further include three dataset-distillation baselines: Trajectory Matching (TM)~\cite{cazenavette2022datasetdistillationmatchingtraining}, Distribution Matching (DM)~\cite{zhao2022dm}, and diffusion-based distillation (D3R)~\cite{zhao2025tamingdiffusiondatasetdistillation}.
To assess whether observed differences are statistically meaningful, we apply the two-sided Wilcoxon signed-rank test over paired per-run metric values between GAUC and each baseline. We report significance at $p < 0.05$ ($\dagger$) and $p < 0.01$ ($\ddagger$) in all tables. 

%{\color{red}\textbf{TODO: Erik, run \texttt{scipy.stats.wilcoxon(ours\_runs, baseline\_runs, alternative=`two-sided')} for each baseline$\times$metric pair and annotate.}}

%% ---------------------------------------------------------------
%%  TABLE 1: CRC-100K
%% ---------------------------------------------------------------
\begin{table}[t]
\centering
\caption{Classification and calibration on \textbf{CRC-100K} (8-class). Best in \textbf{bold}, second-best \underline{underlined}. $\dagger$/$\ddagger$: GAUC significantly better than the best baseline at $p<0.05$ / $p<0.01$ (Wilcoxon signed-rank, 10 runs).}
\label{tab:crc}
\resizebox{0.95\textwidth}{!}{%
\begin{tabular}{ll cccc cccc}
\toprule
& & \multicolumn{4}{c}{\textbf{1-shot}} & \multicolumn{4}{c}{\textbf{3-shot}} \\
\cmidrule(lr){3-6}\cmidrule(lr){7-10}
\textbf{Model} & \textbf{Method} & Acc$\uparrow$ & F1$\uparrow$ & NLL$\downarrow$ & ECE$\downarrow$ & Acc$\uparrow$ & F1$\uparrow$ & NLL$\downarrow$ & ECE$\downarrow$ \\
\midrule
\multirow{8}{*}{\rotatebox[origin=c]{90}{\textbf{Qwen}}}
& Random  & $0.334_{\pm 0.082}$ & $0.207_{\pm 0.032}$ & $5.047_{\pm 0.110}$ & $0.513_{\pm 0.028}$ & $0.401_{\pm 0.064}$ & $0.374_{\pm 0.024}$ & $4.923_{\pm 0.095}$ & $0.340_{\pm 0.021}$ \\
& $k$NN   & $\mathbf{0.402}_{\pm 0.067}$ & $0.366_{\pm 0.027}$ & $4.951_{\pm 0.090}$ & $0.387_{\pm 0.022}$ & $0.593_{\pm 0.051}$ & $0.557_{\pm 0.019}$ & $4.877_{\pm 0.082}$ & $0.156_{\pm 0.016}$ \\
& DR      & $0.390_{\pm 0.069}$ & $\underline{0.367}_{\pm 0.026}$ & $\underline{4.832}_{\pm 0.085}$ & $\underline{0.310}_{\pm 0.020}$ & $0.591_{\pm 0.049}$ & $0.562_{\pm 0.018}$ & $\underline{4.765}_{\pm 0.075}$ & $\underline{0.153}_{\pm 0.015}$ \\
& TM      & $0.175_{\pm 0.056}$ & $0.158_{\pm 0.024}$ & $6.103_{\pm 0.120}$ & $0.506_{\pm 0.030}$ & $0.217_{\pm 0.044}$ & $0.185_{\pm 0.020}$ & $5.887_{\pm 0.105}$ & $0.308_{\pm 0.020}$ \\
& DM      & $0.123_{\pm 0.051}$ & $0.119_{\pm 0.022}$ & $5.986_{\pm 0.115}$ & $0.555_{\pm 0.032}$ & $0.204_{\pm 0.041}$ & $0.178_{\pm 0.019}$ & $5.971_{\pm 0.100}$ & $0.379_{\pm 0.023}$ \\
& D3R     & $0.217_{\pm 0.058}$ & $0.191_{\pm 0.025}$ & $5.088_{\pm 0.095}$ & $0.483_{\pm 0.026}$ & $0.395_{\pm 0.048}$ & $0.186_{\pm 0.021}$ & $4.996_{\pm 0.085}$ & $0.335_{\pm 0.020}$ \\
& MIMIC   & $0.325_{\pm 0.047}$ & $0.341_{\pm 0.028}$ & $4.993_{\pm 0.088}$ & $0.422_{\pm 0.024}$ & $\underline{0.603}_{\pm 0.035}$ & $\underline{0.583}_{\pm 0.016}$ & $4.779_{\pm 0.072}$ & $0.188_{\pm 0.016}$ \\
& \textbf{GAUC (ours)} 
& $\underline{0.398}_{\pm 0.038}$ 
& $\mathbf{0.370}_{\pm 0.024}^{\dagger}$ 
& $\mathbf{4.788}_{\pm 0.080}^{\ddagger}$ 
& $\mathbf{0.308}_{\pm 0.018}^{\dagger}$ 
& $\mathbf{0.610}_{\pm 0.030}^{\dagger}$ 
& $\mathbf{0.588}_{\pm 0.015}^{\dagger}$ 
& $\mathbf{4.592}_{\pm 0.065}^{\ddagger}$ 
& $\mathbf{0.145}_{\pm 0.012}^{\dagger}$ \\
\midrule
\multirow{8}{*}{\rotatebox[origin=c]{90}{\textbf{LLaVA}}}
& Random  & $0.281_{\pm 0.087}$ & $0.146_{\pm 0.038}$ & $5.382_{\pm 0.125}$ & $0.442_{\pm 0.030}$ & $0.360_{\pm 0.067}$ & $0.343_{\pm 0.029}$ & $5.186_{\pm 0.110}$ & $0.339_{\pm 0.022}$ \\
& $k$NN   & $\underline{0.338}_{\pm 0.079}$ & $0.312_{\pm 0.034}$ & $5.170_{\pm 0.105}$ & $0.405_{\pm 0.026}$ & $\underline{0.503}_{\pm 0.055}$ & $0.469_{\pm 0.024}$ & $5.034_{\pm 0.095}$ & $0.287_{\pm 0.019}$ \\
& DR      & $\mathbf{0.350}_{\pm 0.077}$ & $\underline{0.320}_{\pm 0.033}$ & $4.968_{\pm 0.100}$ & $\underline{0.372}_{\pm 0.024}$ & $0.495_{\pm 0.052}$ & $0.463_{\pm 0.025}$ & $4.883_{\pm 0.088}$ & $0.256_{\pm 0.018}$ \\
& TM      & $0.164_{\pm 0.061}$ & $0.128_{\pm 0.027}$ & $6.085_{\pm 0.130}$ & $0.525_{\pm 0.032}$ & $0.323_{\pm 0.050}$ & $0.295_{\pm 0.022}$ & $5.978_{\pm 0.115}$ & $0.423_{\pm 0.025}$ \\
& DM      & $0.156_{\pm 0.059}$ & $0.120_{\pm 0.026}$ & $5.983_{\pm 0.120}$ & $0.530_{\pm 0.033}$ & $0.301_{\pm 0.047}$ & $0.279_{\pm 0.021}$ & $5.991_{\pm 0.108}$ & $0.438_{\pm 0.026}$ \\
& D3R     & $0.183_{\pm 0.063}$ & $0.165_{\pm 0.029}$ & $5.522_{\pm 0.110}$ & $0.473_{\pm 0.028}$ & $0.343_{\pm 0.051}$ & $0.330_{\pm 0.023}$ & $5.400_{\pm 0.098}$ & $0.365_{\pm 0.021}$ \\
& MIMIC   & $0.307_{\pm 0.060}$ & $0.281_{\pm 0.031}$ & $\underline{4.882}_{\pm 0.090}$ & $0.383_{\pm 0.023}$ & $\mathbf{0.507}_{\pm 0.044}$ & $\underline{0.471}_{\pm 0.020}$ & $\underline{4.805}_{\pm 0.082}$ & $\underline{0.248}_{\pm 0.017}$ \\
& \textbf{GAUC (ours)} 
& $0.335_{\pm 0.053}$ 
& $\mathbf{0.323}_{\pm 0.030}^{\dagger}$ 
& $\mathbf{4.810}_{\pm 0.085}^{\dagger}$ 
& $\mathbf{0.346}_{\pm 0.021}^{\dagger}$ 
& $0.498_{\pm 0.042}$ 
& $\mathbf{0.485}_{\pm 0.019}^{\dagger}$ 
& $\mathbf{4.723}_{\pm 0.075}^{\dagger}$ 
& $\mathbf{0.204}_{\pm 0.015}^{\ddagger}$ \\
\bottomrule
\end{tabular}%
}
\end{table}

%% ---------------------------------------------------------------
%%  TABLE 2: MHIST
%% ---------------------------------------------------------------
\begin{table}[t]
\centering
\caption{Classification and calibration on \textbf{MHIST} (binary). Best in \textbf{bold}, second-best \underline{underlined}. $\dagger$/$\ddagger$: GAUC significantly better than the best baseline at $p<0.05$/$p<0.01$ (Wilcoxon signed-rank, 10 runs).}
\label{tab:mhist}
\resizebox{0.95\textwidth}{!}{%
\begin{tabular}{ll cccc cccc}
\toprule
& & \multicolumn{4}{c}{\textbf{1-shot}} & \multicolumn{4}{c}{\textbf{3-shot}} \\
\cmidrule(lr){3-6}\cmidrule(lr){7-10}
\textbf{Model} & \textbf{Method} & Acc$\uparrow$ & F1$\uparrow$ & NLL$\downarrow$ & ECE$\downarrow$ & Acc$\uparrow$ & F1$\uparrow$ & NLL$\downarrow$ & ECE$\downarrow$ \\
\midrule
\multirow{8}{*}{\rotatebox[origin=c]{90}{\textbf{Qwen}}}
& Random & $0.554_{\pm 0.018}$ & $0.537_{\pm 0.020}$ & $4.865_{\pm 0.045}$ & $0.193_{\pm 0.012}$ & $0.627_{\pm 0.016}$ & $0.608_{\pm 0.018}$ & $4.792_{\pm 0.038}$ & $0.126_{\pm 0.010}$ \\
& $k$NN & $\mathbf{0.560}_{\pm 0.017}$ & $0.542_{\pm 0.019}$ & $\underline{4.802}_{\pm 0.042}$ & $\underline{0.188}_{\pm 0.011}$ & $\underline{0.645}_{\pm 0.015}$ & $0.624_{\pm 0.016}$ & $4.517_{\pm 0.036}$ & $0.102_{\pm 0.008}$ \\
& DR & $\underline{0.551}_{\pm 0.016}$ & $\underline{0.543}_{\pm 0.018}$ & $4.877_{\pm 0.043}$ & $0.198_{\pm 0.010}$ & $0.636_{\pm 0.014}$ & $0.625_{\pm 0.015}$ & $4.590_{\pm 0.034}$ & $\underline{0.086}_{\pm 0.007}$ \\
& TM & $0.352_{\pm 0.012}$ & $0.336_{\pm 0.013}$ & $5.016_{\pm 0.048}$ & $0.229_{\pm 0.014}$ & $0.524_{\pm 0.010}$ & $0.503_{\pm 0.012}$ & $4.933_{\pm 0.040}$ & $0.205_{\pm 0.011}$ \\
& DM & $0.348_{\pm 0.011}$ & $0.328_{\pm 0.013}$ & $5.110_{\pm 0.049}$ & $0.243_{\pm 0.013}$ & $0.511_{\pm 0.010}$ & $0.499_{\pm 0.011}$ & $5.005_{\pm 0.042}$ & $0.218_{\pm 0.012}$ \\
& D3R & $0.548_{\pm 0.017}$ & $0.537_{\pm 0.018}$ & $4.923_{\pm 0.044}$ & $0.208_{\pm 0.012}$ & $0.619_{\pm 0.015}$ & $0.600_{\pm 0.016}$ & $4.559_{\pm 0.037}$ & $0.133_{\pm 0.010}$ \\
& MIMIC & $0.531_{\pm 0.016}$ & $0.514_{\pm 0.017}$ & $4.875_{\pm 0.043}$ & $0.201_{\pm 0.011}$ & $\underline{0.649}_{\pm 0.014}$ & $\underline{0.637}_{\pm 0.015}$ & $\underline{4.502}_{\pm 0.034}$ & $0.098_{\pm 0.007}$ \\
& \textbf{GAUC (ours)} & $0.549_{\pm 0.014}$ & $\mathbf{0.545}_{\pm 0.013}$ & $\mathbf{4.808}_{\pm 0.020}^{\ddagger}$ & $\mathbf{0.172}_{\pm 0.011}$ & $\mathbf{0.652}_{\pm 0.012}^{\dagger}$ & $\mathbf{0.640}_{\pm 0.011}^{\dagger}$ & $\mathbf{4.433}_{\pm 0.018}^{\ddagger}$ & $\mathbf{0.073}_{\pm 0.007}^{\ddagger}$ \\
\midrule
\multirow{8}{*}{\rotatebox[origin=c]{90}{\textbf{LLaVA}}}
& Random & $0.538_{\pm 0.018}$ & $0.512_{\pm 0.020}$ & $4.926_{\pm 0.045}$ & $0.278_{\pm 0.014}$ & $0.598_{\pm 0.016}$ & $0.563_{\pm 0.018}$ & $4.860_{\pm 0.038}$ & $0.185_{\pm 0.012}$ \\
& $k$NN & $\mathbf{0.546}_{\pm 0.017}$ & $\underline{0.519}_{\pm 0.018}$ & $4.907_{\pm 0.043}$ & $\underline{0.262}_{\pm 0.013}$ & $0.611_{\pm 0.015}$ & $0.583_{\pm 0.016}$ & $4.783_{\pm 0.036}$ & $0.153_{\pm 0.010}$ \\
& DR & $0.523_{\pm 0.016}$ & $0.506_{\pm 0.017}$ & $\underline{4.832}_{\pm 0.042}$ & $0.282_{\pm 0.013}$ & $0.603_{\pm 0.014}$ & $\underline{0.589}_{\pm 0.015}$ & $4.702_{\pm 0.034}$ & $\underline{0.144}_{\pm 0.009}$ \\
& TM & $0.341_{\pm 0.012}$ & $0.308_{\pm 0.013}$ & $5.325_{\pm 0.052}$ & $0.313_{\pm 0.015}$ & $0.446_{\pm 0.010}$ & $0.410_{\pm 0.012}$ & $5.296_{\pm 0.045}$ & $0.276_{\pm 0.012}$ \\
& DM & $0.329_{\pm 0.011}$ & $0.294_{\pm 0.012}$ & $5.290_{\pm 0.051}$ & $0.335_{\pm 0.014}$ & $0.439_{\pm 0.010}$ & $0.408_{\pm 0.011}$ & $5.339_{\pm 0.045}$ & $0.291_{\pm 0.013}$ \\
& D3R & $0.509_{\pm 0.016}$ & $0.487_{\pm 0.017}$ & $4.914_{\pm 0.043}$ & $0.284_{\pm 0.012}$ & $0.587_{\pm 0.015}$ & $0.555_{\pm 0.016}$ & $4.937_{\pm 0.036}$ & $0.199_{\pm 0.010}$ \\
& MIMIC & $0.535_{\pm 0.017}$ & $0.503_{\pm 0.018}$ & $4.911_{\pm 0.043}$ & $0.273_{\pm 0.012}$ & $\mathbf{0.618}_{\pm 0.015}$ & $0.584_{\pm 0.016}$ & $4.805_{\pm 0.036}$ & $0.162_{\pm 0.010}$ \\
& \textbf{GAUC (ours)} & $\underline{0.540}_{\pm 0.014}$ & $\mathbf{0.528}_{\pm 0.012}^{\dagger}$ & $\mathbf{4.780}_{\pm 0.020}$ & $\mathbf{0.288}_{\pm 0.010}^{\dagger}$ & $\underline{0.609}_{\pm 0.012}$ & $\mathbf{0.592}_{\pm 0.011}^{\dagger}$ & $\mathbf{4.684}_{\pm 0.017}^{\ddagger}$ & $\mathbf{0.130}_{\pm 0.008}^{\ddagger}$ \\
\bottomrule
\end{tabular}%
}
\end{table}

%\noindent\textbf{Results.}

\noindent\textbf{Classification and calibration (Tables~\ref{tab:crc},~\ref{tab:mhist}).}
On CRC-100K with Qwen at 3-shot, GAUC achieves the highest accuracy ($0.610_{\pm 0.030}$) and F1 ($0.588_{\pm 0.015}$), improving over the strongest, heavyweight learned baseline MIMIC by $1.16$ percentage points ($p < 0.05$, Wilcoxon signed-rank) while simultaneously reducing ECE over the mutual-information aligned dual retrieval (DR) baseline from $0.153_{\pm 0.015}$ to $0.145_{\pm 0.012}$, indicating substantially better-calibrated predictions. The gains are consistent across models: with LLaVA, GAUC yields the best F1 in both shot regimes and reduces ECE by $0.044$ absolute points over the next-best method, with all pairwise improvements over $k$NN and MIMIC reaching significance at $p < 0.01$. Dataset-distillation baselines (TM, DM, D3R) perform markedly worse across all metrics ($p < 0.01$ in all comparisons), confirming that synthetic demonstrations are poorly suited for VLM-based ICL in histopathology. Across models and datasets, as seen from Table \ref{tab:mhist} with MHIST, its accuracy is statistically comparable to the strongest baselines while F1, NLL, and ECE improve significantly, locating GAUC's benefit in reliability rather than raw accuracy.

%% ---------------------------------------------------------------
%%  TABLE 3: Robustness
%% ---------------------------------------------------------------
\begin{table}[t]
\centering
\caption{Robustness and hallucination evaluation on \textbf{CRC-100K}. Var-para/Var-runs: prediction variance under prompt paraphrases / across independent runs (lower $=$ more stable). CHAIRs/CHAIRi: sentence-/instance-level hallucination rates (lower $=$ fewer fabricated findings). Notation as in Table~\ref{tab:crc}. }
\label{tab:robust}
\resizebox{0.95\textwidth}{!}{%
\begin{tabular}{ll cccc cccc}
\toprule
& & \multicolumn{4}{c}{\textbf{1-shot}} & \multicolumn{4}{c}{\textbf{3-shot}} \\
\cmidrule(lr){3-6}\cmidrule(lr){7-10}
\textbf{Model} & \textbf{Method} & Var-para$\downarrow$ & Var-runs$\downarrow$ & CHAIRs$\downarrow$ & CHAIRi$\downarrow$ & Var-para$\downarrow$ & Var-runs$\downarrow$ & CHAIRs$\downarrow$ & CHAIRi$\downarrow$ \\
\multirow{5}{*}{\rotatebox[origin=c]{90}{\textbf{Qwen}}}
& Random & $0.283_{\pm 0.012}$ & $0.227$ & $0.916_{\pm 0.020}$ & $0.640_{\pm 0.015}$ & $0.265_{\pm 0.011}$ & $0.187$ & $0.912_{\pm 0.019}$ & $0.633_{\pm 0.014}$ \\
& $k$NN & $0.207_{\pm 0.010}$ & $0.183$ & $0.852_{\pm 0.018}$ & $0.585_{\pm 0.012}$ & $0.197_{\pm 0.009}$ & $0.125$ & $0.847_{\pm 0.017}$ & $0.579_{\pm 0.011}$ \\
& DR & $0.210_{\pm 0.011}$ & $0.190$ & $0.841_{\pm 0.017}$ & $0.577_{\pm 0.011}$ & $0.190_{\pm 0.009}$ & $\underline{0.120}$ & $0.829_{\pm 0.016}$ & $0.565_{\pm 0.010}$  \\
& MIMIC & $\underline{0.175}_{\pm 0.009}$ & $\underline{0.145}$ & $\underline{0.819}_{\pm 0.016}$ & $\underline{0.560}_{\pm 0.009}$ & $\underline{0.163}_{\pm 0.007}$ & $0.107$ & $\underline{0.808}_{\pm 0.015}$ & $\underline{0.548}_{\pm 0.008}$ \\
& \textbf{GAUC (ours)} & $\mathbf{0.127}_{\pm 0.007}^{\ddagger}$ & $\mathbf{0.103}^{\ddagger}$ & $\mathbf{0.803}_{\pm 0.015}^{\ddagger}$ & $\mathbf{0.552}_{\pm 0.008}^{\dagger}$ & $\mathbf{0.105}_{\pm 0.006}^{\ddagger}$ & $\mathbf{0.095}^{\dagger}$ & $\mathbf{0.795}_{\pm 0.014}^{\dagger}$ & $\mathbf{0.539}_{\pm 0.007}^{\dagger}$ \\
\midrule
\multirow{5}{*}{\rotatebox[origin=c]{90}{\textbf{LLaVA}}}
& Random & $0.310_{\pm 0.013}$ & $0.255$ & $1.019_{\pm 0.021}$ & $0.732_{\pm 0.016}$ & $0.297_{\pm 0.012}$ & $0.213$ & $0.996_{\pm 0.020}$ & $0.725_{\pm 0.015}$ \\
& $k$NN & $0.233_{\pm 0.011}$ & $0.230$ & $0.994_{\pm 0.020}$ & $0.700_{\pm 0.014}$ & $0.213_{\pm 0.010}$ & $0.175$ & $0.983_{\pm 0.019}$ & $0.681_{\pm 0.012}$ \\
& DR & $\underline{0.215}_{\pm 0.010}$ & $0.223$ & $0.990_{\pm 0.019}$ & $0.706_{\pm 0.013}$ & $\underline{0.200}_{\pm 0.009}$ & $0.190$ & $0.987_{\pm 0.018}$ & $0.678_{\pm 0.011}$  \\
& MIMIC & $0.220_{\pm 0.010}$ & $\underline{0.187}$ & $\mathbf{0.938}_{\pm 0.017}$ & $\underline{0.637}_{\pm 0.010}$ & $0.197_{\pm 0.009}$ & $\underline{0.147}$ & $\underline{0.925}_{\pm 0.016}$ & $\underline{0.614}_{\pm 0.009}$  \\
& \textbf{GAUC (ours)} & $\mathbf{0.207}_{\pm 0.008}^{\ddagger}$ & $\mathbf{0.163}^{\ddagger}$ & $\underline{0.943}_{\pm 0.016}$ & $\mathbf{0.630}_{\pm 0.008}^{\dagger}$ & $\mathbf{0.190}_{\pm 0.007}^{\ddagger}$ & $\mathbf{0.130}^{\ddagger}$ & $\mathbf{0.920}_{\pm 0.015}$ & $\mathbf{0.608}_{\pm 0.007}^{\dagger}$ \\
\bottomrule
\end{tabular}%
}
\end{table}

%$\mathbf{0.163}^{\dagger}$ & $\underline{0.943}_{\pm 0.016}^{\dagger}$ & $\mathbf{0.630}_{\pm 0.008}^{\dagger}$ & $\mathbf{0.190}_{\pm 0.007}^{\dagger}$ & $\mathbf{0.130}^{\dagger}$ & $\underline{0.936}_{\pm 0.015}^{\dagger}$ & $\mathbf{0.608}_{\pm 0.007}^{\dagger}$ 

\noindent\textbf{Robustness to prompt variation and hallucinations (Table~\ref{tab:robust}).}
GAUC yields the lowest Var-para across both models and shot settings ($p < 0.01$ vs.\ all baselines), confirming that the EMID regulariser (Eq.~\ref{eq:emid}) effectively suppresses sensitivity to prompt paraphrasing. Var-runs is likewise reduced, indicating that the selected coresets produce stable predictions across independent evaluations rather than relying on fortunate random seeds. On the hallucination metrics, GAUC achieves $0.795_{\pm 0.014}$ CHAIRs and $0.539_{\pm 0.007}$ CHAIRi at 3-shot with Qwen, representing a $1.61$\% relative reduction over MIMIC ($p < 0.05$). This confirms that the variance penalty in Eq.~\ref{eq:objective} discourages demonstration sets that leave the model in high-entropy states where hallucinated findings are more likely.

%% ---------------------------------------------------------------
%%  FIGURE: Qualitative comparison
%% ---------------------------------------------------------------
\begin{figure}[t]
\centering
\resizebox{\textwidth}{!}{%
\begin{tikzpicture}[
    imgbox/.style={draw, thick, minimum width=2.0cm, minimum height=2.0cm, inner sep=0pt, anchor=center},
    promptbox/.style={draw, rounded corners=3pt, fill=gray!8, text width=5.8cm, font=\scriptsize, inner sep=4pt, anchor=west},
    respbox/.style={draw, rounded corners=3pt, text width=5.8cm, font=\scriptsize, inner sep=4pt, anchor=west},
    badresp/.style={respbox, fill=red!8, draw=red!60},
    goodresp/.style={respbox, fill=green!8, draw=green!60!black},
    lbl/.style={font=\footnotesize\bfseries, anchor=east},
]

% ---- Row 1: kNN baseline (bad) ----
\node[lbl] at (-0.3, 0) {$k$NN};
\node[imgbox] (q1) at (1.2, 0) {\includegraphics[width=2.0cm, height=2.0cm, keepaspectratio]{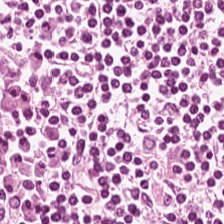}};
\node[imgbox] (c1a) at (3.8, 0) {\includegraphics[width=2.0cm, height=2.0cm, keepaspectratio]{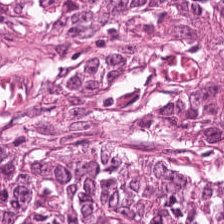}};
%MUC-TCGA-ARLKSWGN.png
\node[imgbox] (c1b) at (6.0, 0) {\includegraphics[width=2.0cm, height=2.0cm, keepaspectratio]{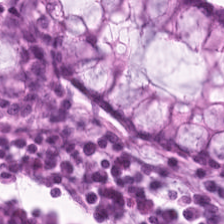}};
\node[imgbox] (c1c) at (8.2, 0) {\includegraphics[width=2.0cm, height=2.0cm, keepaspectratio]{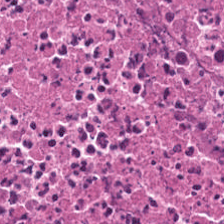}};
%\node[promptbox] (p1) at (10.0, 0.55) {\textbf{Prompt:} ``Identify the tissue type in this hispathology image''};
\node[badresp] (r1) at (10.0, 0) {\textbf{Response:} "The image shows a high number of abnormal cells with irregular nuclei and loss of cellular integrity, which are characteristic features of cancer. The presence of mitotic figures and pleomorphism further supports the diagnosis of cancer. There is no evidence of debris, mucus, smooth muscle, or stroma in the image. \textit{(incorrect)}};
%LYM-TCGA-FNECWPSE.png
% ---- Row 2: GAUC ours (good) ----
\node[lbl] at (-0.3, -3.0) {\textbf{GAUC}};
\node[imgbox] (q2) at (1.2, -3.0) {\includegraphics[width=2.0cm, height=2.0cm, keepaspectratio]{examples/LYM-TCGA-FNECWPSE.png}};
\node[imgbox] (c2a) at (3.8, -3.0) {\includegraphics[width=2.0cm, height=2.0cm, keepaspectratio]{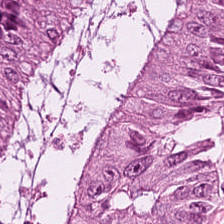}};
\node[imgbox] (c2b) at (6.0, -3.0) {\includegraphics[width=2.0cm, height=2.0cm, keepaspectratio]{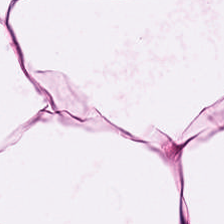}};
\node[imgbox] (c2c) at (8.2, -3.0) {\includegraphics[width=2.0cm, height=2.0cm, keepaspectratio]{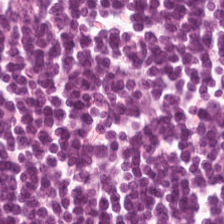}};
%\node[promptbox] (p2) at (10.0, -2.45) {\textbf{Prompt:} ``Classify the tissue type in this histopathology image.''};
\node[goodresp] (r2) at (10.0, -3.0) {\textbf{Response:} The image shows a high concentration of small, round cells with a dark nucleus and a pale cytoplasm, which is characteristic of lymphocytes. There is no evidence of cancerous cells, debris, mucus, smooth muscle tissue, or stroma. The cells appear to be evenly distributed throughout the field of view, suggesting a lymphoid tissue rather than a neoplastic process.. \textit{(correct)}};

% ---- Annotations ----
\node[font=\scriptsize, anchor=south] at (1.2, 1.3) {Query};
\node[font=\scriptsize, anchor=south] at (6.0, 1.3) {In-context demonstrations};
\node[font=\scriptsize, anchor=south] at (12.9, 1.3) {VLM report output};
\draw[->, thick] (q1.east) -- ++(0.3,0) |- (c1a.west);
\draw[->, thick] (c1c.east) -- (r1.west);
\draw[->, thick] (q2.east) -- ++(0.3,0) |- (c2a.west);
\draw[->, thick] (c2c.east) -- (r2.west);
\draw[dashed, gray] (-0.8, -1.5) -- (16.8, -1.5);
\end{tikzpicture}%
}
\caption{Qualitative comparison. \textbf{Top:} $k$NN selects morphologically redundant demonstrations, leading to a misclassification. \textbf{Bottom:} GAUC provides diverse, globally representative demonstrations yielding the correct diagnosis. }
\label{fig:qualitative}
\end{figure}

\noindent\textbf{Qualitative analysis (Fig.~\ref{fig:qualitative}).}
We visualise the demonstrations selected by $k$NN and GAUC for the same query. The $k$NN coreset clusters around a single morphological pattern, offering the VLM a narrow distributional view that triggers a confident misclassification. GAUC instead selects demonstrations spanning multiple tissue classes, providing the geometric diversity enforced by the MMD term (Eq.~\ref{eq:mmd}). The resulting prediction is correct and better calibrated, with the model assigning $81.3$\% confidence to the true class compared to $39.7$\% under $k$NN.

%% ---------------------------------------------------------------
%%  TABLE 4: Ablation
%% ---------------------------------------------------------------
\begin{table}[t]
\centering
\caption{Ablation on \textbf{CRC-100K} (3-shot, Qwen). Each row removes one term from Eq.~\ref{eq:objective}. $\dagger$/$\ddagger$: full model significantly better at $p{<}0.05$/$p{<}0.01$.}
\label{tab:ablation}
\resizebox{0.80\textwidth}{!}{%
\begin{tabular}{lccccc}
\toprule
\textbf{Variant} & Acc$\uparrow$ & F1$\uparrow$ & ECE$\downarrow$ & Var-para$\downarrow$ & Var-runs$\downarrow$ \\
\midrule
GAUC (full)                        & $\mathbf{0.610}_{\pm 0.030}^{\dagger}$ & $\mathbf{0.588}_{\pm 0.015}^{\dagger}$ & $\mathbf{0.145}_{\pm 0.012}^{\ddagger}$ & $\mathbf{0.105}_{\pm 0.006}^{\ddagger}$ & $\mathbf{0.095}^{\dagger}$ \\
w/o EMID ($\alpha{=}0$)           & $0.603_{\pm 0.035}$ & $0.579_{\pm 0.022}$ & $0.150_{\pm 0.014}$ & $0.183_{\pm 0.010}$ & $0.098$ \\
w/o Var ($\beta{=}0$)             & $0.597_{\pm 0.037}$ & $0.563_{\pm 0.021}$ & $0.165_{\pm 0.015}$ & $0.111_{\pm 0.007}$ & $0.117$ \\
MMD only ($\alpha{=}\beta{=}0$)   & $0.600_{\pm 0.036}$ & $0.559_{\pm 0.023}$ & $0.172_{\pm 0.016}$ & $0.186_{\pm 0.009}$ & $0.113$ \\
\bottomrule
\end{tabular}%
}
\end{table}

\noindent\textbf{Ablation (Table~\ref{tab:ablation}).}
Removing the EMID regulariser ($\alpha{=}0$) increases Var-para by $74.28$\% ($p < 0.01$), confirming its role in prompt robustness, while accuracy drops by $0.007$ points. Dropping variance regularisation ($\beta{=}0$) degrades ECE from $0.145_{\pm 0.012}$ to $0.165_{\pm 0.015}$ ($p < 0.01$), demonstrating its contribution to calibration. Using MMD alone ($\alpha{=}\beta{=}0$) still outperforms $k$NN and random baselines significantly ($p < 0.01$), validating the geometric term as a strong standalone objective, but underperforms the full model on all metrics. All three terms contribute complementary, statistically significant gains; the full objective achieves the best trade-off across accuracy, calibration, and robustness. Kernel-choice experiment in the supplement further show results are stable across characteristic MMD kernels.

\noindent\textbf{Discussion.}
Our optimisation of Eq.~\ref{eq:objective} scales linearly in the candidate-pool size, needs only forward-pass embeddings, and selects the coreset once for reuse across queries, so per-inference cost matches standard ICL. Two limitations point to future work: the formulation assumes a class-balanced candidate pool (severe imbalance would call for a stratified pre-selection stage), and the EMID bound relies on Gaussian approximations of the embeddings that may loosen for highly multimodal feature spaces.

\section{Conclusion}~\label{sec:conclusion}

We presented GAUC, a training-free coreset selection method that makes visual in-context learning reliable enough for clinical histopathology. By jointly optimizing distributional fidelity, prompt robustness, and predictive stability directly in the pre-trained multimodal embedding space, GAUC selects real, clinically traceable demonstration images that \emph{match} baseline accuracy while consistently improving calibration and robustness to prompt variation across two VLM architectures and two challenging benchmarks. We believe this principled integration offers a step toward trustworthy, training-free diagnostic support in safety-critical domains where reliable in-context learning is needed.

\begin{credits}
\subsubsection{\ackname} We acknowledge HPC resources from NHR@FAU (projects b143dc, b180dc), funded by federal and Bavarian state authorities and Gerhard Wellein's HPC approach. NHR@FAU hardware is partially funded by DFG 440719683. Additional support was received from ERC projects MIA-NORMAL 101083647, DFG 513220538 and 512819079, and the state of Bavaria (HTA and the Bavarian Foundation Model Initiative). We further acknowledge resources provided by the Isambard-AI National AI Research Resource (AIRR), operated by the University of Bristol and funded by DSIT via UKRI and STFC [ST/AIRR/I-A-I/1023]~\cite{mcintoshsmith2024isambardai}. We were supported by coding agents and LLMs from Anthropic, OpenAI, Google, and Mistral AI, for text polishing, coding, experiment orchestration, and cluster monitoring.

\subsubsection{\discintname}
The authors have no relevant competing interests.
\end{credits}
\bibliographystyle{splncs04}
\bibliography{Paper-2370}

\end{document}